\documentclass[10pt,twocolumn,letterpaper]{article}

\usepackage[pagenumbers]{cvpr} 

\definecolor{cvprblue}{rgb}{0.21,0.49,0.74}
\usepackage[pagebackref,breaklinks,colorlinks,allcolors=cvprblue]{hyperref}
\usepackage{multirow}

\usepackage{listings}
\def\paperID{*****} 
\def\confName{CVPR}
\def\confYear{2025}

\title{How Visual Representations Map to Language Feature Space in Multimodal LLMs}

\author{
Constantin Venhoff\\
University of Oxford
\and
Ashkan Khakzar\\
University of Oxford
\and
Sonia Joseph\\
McGill University / Meta
\and
Philip Torr\\
University of Oxford
\and
Neel Nanda
}

\begin{document}
\maketitle
\begin{abstract}
Effective multimodal reasoning depends on the alignment of visual and linguistic representations, yet the mechanisms by which vision-language models (VLMs) achieve this alignment remain poorly understood. Following the LiMBeR framework, we deliberately maintain a frozen large language model (LLM) and a frozen vision transformer (ViT), connected solely by training a linear adapter during visual instruction tuning. By keeping the language model frozen, we ensure it maintains its original language representations without adaptation to visual data. Consequently, the linear adapter must map visual features directly into the LLM's existing representational space rather than allowing the language model to develop specialized visual understanding through fine-tuning.
Our experimental design uniquely enables the use of pre-trained sparse autoencoders (SAEs) of the LLM as analytical probes. These SAEs remain perfectly aligned with the unchanged language model and serve as a snapshot of the learned language feature-representations. Through systematic analysis of SAE reconstruction error, sparsity patterns, and feature SAE descriptions, we reveal the layer-wise progression through which visual representations gradually align with language feature representations, converging in middle-to-later layers. This suggests a fundamental misalignment between ViT outputs and early LLM layers, raising important questions about whether current adapter-based architectures optimally facilitate cross-modal representation learning.
\footnote{Trained weights and code available at \url{https://github.com/cvenhoff/vlm-mapping}}
\end{abstract}    
\section{Introduction}
\begin{table*}[ht]
    \centering
    \small
    \begin{tabular}{l l c c c c}
        \hline
        \textbf{Task} & \textbf{Metric} & \textbf{CLIP-Gemma-2-2B-it} & \textbf{LLaVA-v1-7B}\cite{liu2023visual} & \textbf{LLaVA-v1.5-7B}\cite{llava15} & \textbf{InstructBlip-7B}\cite{dai2023instructblip} \\
        \hline
        \multirow{2}{*}{MME \cite{mme}}  
            & Cognition   & 291.43  & 243.6 & 302.1 & 254.3 \\
            & Perception  & 920.19  & 832 & 1506.2 & 1137.1 \\
        \hline
        \multirow{4}{*}{LLaVA Bench \cite{liu2023visual}} 
            & Conv  & 63.7 & 54.7 & 54.4 & 69.3 \\
            & Complex & 52.5  & 66.1 & 70.5 & 59.3 \\
            & Detail & 35.4  & 44.2 & 55 & 48.2 \\
            & All & 50.3  & 57.2 & 61.8 & 59.8 \\
        \hline
        GQA \cite{gqa} & Exact Match & 39.58 & - & 62.0 & 49.5 \\
        \hline
        \multirow{4}{*}{POPE \cite{pope}} 
            & Acc  & 72.57  & 69.2 & 87 & 86 \\
            & Prec & 68.1  & 62.3 & 92.1 & 85.7 \\
            & Rec  & 84.91  & 96.9 & 80.9 & 86.5 \\
            & F1   & 75.58  & 75.9 & 86.1 & 86.1 \\
        \hline
    \end{tabular}
    \caption{Performance of our experimental model (CLIP-Gemma-2-2B-it) using only a trainable linear adapter between frozen vision and language backbones, compared against established VLMs. Despite this constrained configuration, the model performs comparable to LLaVA-v1-7B (with LLM fine-tuning), while LLaVA-v1.5-7B (with end-to-end fine-tuning) and InstructBLIP (with a vast dataset and a transformer-based projector) serve as upper performance bounds. This validates our approach for investigating cross-modal representation alignment without confounding effects from language model adaptation.}
    \label{tab:performance}

\end{table*}

Vision language models (VLMs) \cite{liu2023visual,lin2024vila,steiner2024paligemma2familyversatile} have emerged as powerful systems capable of processing and reasoning across text and visual inputs. These models typically integrate a vision transformer and a language model via a connector component, enabling them to understand images and answer visual queries. Notably, VLMs can be trained efficiently through visual instruction tuning, requiring far less paired data than the extensive pretraining of their individual components—suggesting a natural alignment between visual and linguistic representations \cite{liu2023visual,tong2024metamorphmultimodalunderstandinggeneration}. While prior work has examined visual information flow within VLMs, the specific mapping of visual to linguistic representations across LLM layers remains an open question—key to both theoretical insight and practical architectural improvements.

To explore this, we apply mechanistic interpretability tools \cite{davies2024cognitiverevolutioninterpretabilityexplaining,sharkey2025open} to analyze VLM internal representations beyond behavioral outputs. Central to our method are sparse autoencoders (SAEs) \cite{bricken2023mono}, which disentangle neural representations into interpretable features. We build on the LiMBeR framework that demonstrates effective feature mapping from image to language space \cite{merullo2023linearly}), by connecting a frozen vision transformer (CLIP ViT-L14 \cite{radford2021learning}) and a frozen LLM (Gemma-2-2b-it \cite{gemmateam2024gemma2improvingopen}) via a trainable linear adapter. By keeping both the ViT and LLM frozen, we enforce a setting where the adapter must align visual inputs to the LLM’s fixed feature space, without altering the language model’s internal structure.

This setup allows us to leverage pre-trained SAEs from GemmaScope \cite{lieberum2024gemma}, originally trained on Gemma-2-2b, to probe feature alignment across all layers. Because the LLM remains unchanged, the SAEs acts as an analytical probe, letting us assess reconstruction error, feature sparsity, and the semantic alignment of SAE-derived feature descriptions with visual input across layers. Unlike logit lens methods \cite{logitlens,neo2024towards}, which detect alignment only in the unembedding space, SAEs offer insight into representational alignment throughout the model's layers.

Our results unveil the middle-to-late layers as the convergence location for the cross-modal mapping by finding that the semantic alignment of SAE feature descriptions and the visual input steeply increases in these layers, while the SAE reconstruction error and sparsity decrease. This reveals that visual representations gradually become represented by precise, sparse language model features, indicating successful cross-modal mapping from vision to language model feature space. Crucially, this goes beyond simple out-of-distribution detection as it reveals the specific language model features that the adapter uses to represent visual information in the middle-to-late layers. 

These findings shed new light on visual-linguistic representation alignment. Despite the adapter mapping into the LLM’s token space, middle-to-late LLM layers appear more naturally aligned with ViT outputs than early ones. This insight into early-layer mismatch opens new directions for optimizing VLM architectures and understanding the structural dynamics of cross-modal integration.
\section{Related Work}
\label{sec:intro}

\subsection{Multimodal Mechanistic Interpretability}
Recent interpretability research has explored vision-language models through various approaches. \cite{chen2023interpreting} and \cite{gandelsman2023interpreting} analyzed CLIP, while \cite{schwettmann2023multimodal} examined multimodal neurons in VLMs. \cite{jiang2024interpreting} investigated VLM responses to hallucinated versus real objects. \cite{palit2023towards} used Causal Mediation Analysis on BLIP, \cite{neo2024towards} projected visual representations onto language vocabulary, while \cite{venhoff2025too} probed for the emergence of visual representations in the language model backbone.

\subsection{Dictionary Learning and Sparse Autoencoders}
Sparse Dictionary Learning (SDL) methods \cite{huben2023sparse,elhage2021mathematical} decompose model activations into interpretable directions. Sparse Autoencoders (SAEs) \cite{bricken2023mono}, a type of SDL, enable comprehensive analysis across all network layers through enforced activation sparsity. To avoid the computational cost of training our own SAEs, we use Gemma Scope \cite{lieberum2024gemma}, pre-trained SAEs for Gemma2-2b \cite{team2024gemma}.

\subsection{Linear Mapping Approaches}
LiMBeR \cite{merullo2023linearly} introduced the setup of mapping frozen vision encoder outputs into a frozen LLM’s input space via a trained linear projection, enabling image captioning without tuning either backbone. \cite{park2024bridging} improves on this with an optimal transport loss aligning visual and text representations to word embeddings. \cite{ebrahimi2024cromecrossmodaladaptersefficient} trains only a gated nonlinear adapter for efficient fusion.

    
\section{Training the Cross-Modal Adapter}
\begin{figure*}[t]
    \centering
    \includegraphics[width=\linewidth]{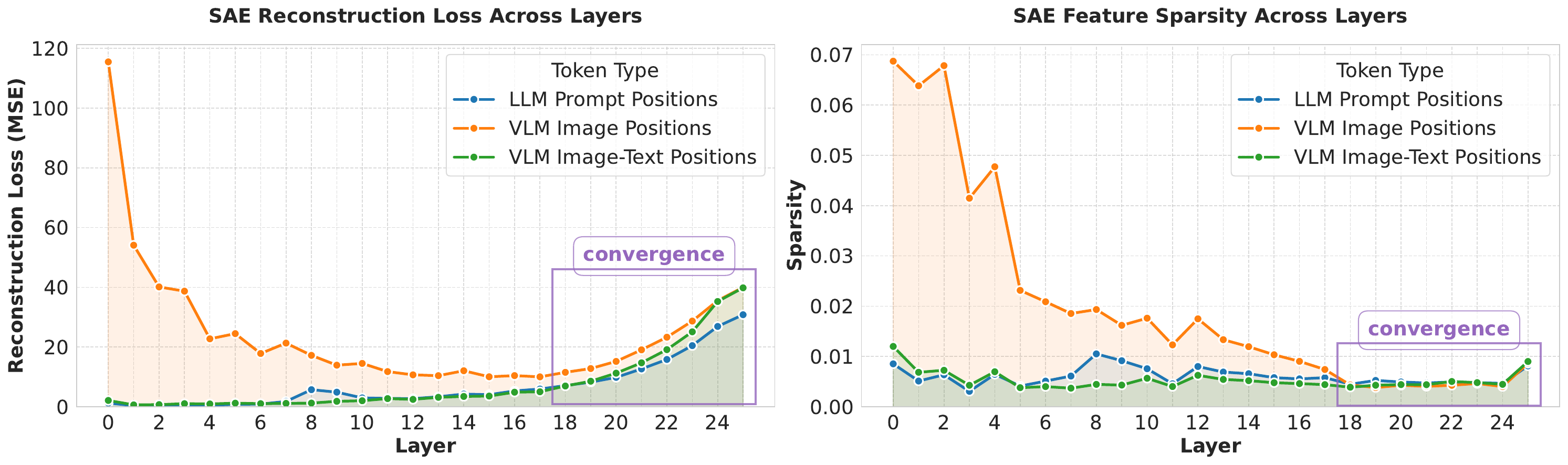}
    \caption{Layer-wise SAE reconstruction error and sparsity patterns. The y-axis shows the MSE reconstruction error (left) and the Sparsity of SAE feature activations as the fraction of non-zero activations (right)."}
    \label{fig:reconsparsity}
\end{figure*}

In this section, we detail the methodology used to train and evaluate our vision language model (VLM). Following the LiMBeR framework \cite{merullo2023linearly} we keep both the vision transformer and the language model frozen during training, ensuring that the learned adapter is solely responsible for mapping visual representations into the existing language feature space.

\subsection{Training Setup}
We employ a LLaVA-style \cite{liu2023visual} adapter training setup, where a vision transformer and a LLM are connected via a learnable linear projection layer. As discussed, we freeze both the vision transformer (ViT) and the large language model (LLM) throughout all training stages. This setup enables precise interpretability of the cross-modal mapping process.

\subsubsection{Dataset}
We utilize the same dataset as LLaVA-1.5 \cite{llava15}, consisting of two primary training stages:

\begin{itemize}
    \item \textbf{Pretraining:} In the first stage a subset of 595K filtered image-text pairs from Conceptual Captions 3M is used. These pairs are formatted into a simple instruction-response style, where the instruction asks for a description of the image, and the response provides the corresponding caption.
    \item \textbf{Finetuning:} In the second stage, a finetuning dataset consisting of 665K instruction-following examples is used. This dataset was introduced by \citeauthor{llava15} for the LLaVA-1.5 finetuning stage.
\end{itemize}

\subsubsection{Model Architecture}
Our VLM uses the CLIP ViT-L/14 as the vision transformer \cite{radford2021learning} and Gemma-2-2b-it, the instruction-following finetuned variant of the Gemma-2-2b model \cite{gemmateam2024gemma2improvingopen}, as the language model backbone. To project the ViT output into the token space of Gemma-2-2b-it we use a linear adapter.

To integrate vision features into the LLM, we process an image through the CLIP ViT encoder and extract 256 visual representations (excluding the CLS token). These tokens are then concatenated after the beginning-of-sequence (BOS) token of the Gemma-2-2b-it model, followed by the remaining language model input tokens.

\subsubsection{Training Hyperparameters}
Our training process is divided into two stages: pretraining and finetuning. During pretraining, we use a batch size of 128, train for one epoch, set the learning rate to $1 \times 10^{-3}$, and apply a warmup ratio of 0.03. In the finetuning stage, we use a batch size of 64, train for three epochs, and set the learning rate to $2 \times 10^{-5}$ without any warmup. Additionally, both the vision transformer (ViT) and the language model (LLM) weights are stored in 16-bit precision.

\subsection{Benchmark Evaluation}

To assess the performance of our trained VLM, we benchmark it against prominent existing VLMs. The evaluation is conducted on standard vision-language benchmarks as presented in Table~\ref{tab:performance}. Our training setup not only maintains a controlled environment for analyzing cross-modal representation learning but also results in a functional and surprisingly solid VLM, despite its small size and few trainable parameters.

\section{Analysis of Cross-Modal Representation Mapping}
To investigate how visual representations integrate into the language model’s feature space, we use sparse autoencoders (SAEs), an interpretability tool that disentangles activations into sparse, interpretable features. Since Gemma-2-2b-it remains frozen during training, we use a GemmaScope SAE \cite{lieberum2024gemma} to analyze how projected visual representations interact with the LLM’s representations. By evaluating the SAE reconstruction error, sparsity patterns, and feature descriptions, we aim to identify the layers where visual and language features optimally align.

\paragraph{Beyond Logit Lens}
While logit lens \cite{logitlens,neo2024towards} may seem viable for this analysis, it has a fundamental limitation: it relies exclusively on the model's unembedding matrix, which projects hidden states onto vocabulary space. This approach cannot capture meaningful representations in layers that do not operate in the unembedding space—this naturally occurs only in later layers. The logit lens thus creates a blind spot precisely where cross-modal integration begins. In contrast, SAEs represent the features of the LLM at every layer and serve as a snapshot of the learned features. This allows us to detect whether visual representations align with language features even in early layers.

\subsection{SAE-Based Analysis Framework}
We first assess the SAE reconstruction error, which measures how well visual representations can be reconstructed based on the language model feature distribution. For each layer $l$, we compute the average reconstruction error across all visual token positions:

\begin{equation}
E_l = \frac{1}{N} \sum_{i=1}^{N} ||\mathbf{V}_l^i - \text{SAE}_l(\mathbf{V}_l^i)||_2^2
\end{equation}

where $N$ is the number of visual tokens and $\text{SAE}_l$ represents the encode-decode operation of the SAE at layer $l$. A steep decline in error suggests a transition point where visual information integrates effectively into the LLM’s representation space.

To ensure this transition is meaningful, we analyze SAE sparsity patterns, measuring the fraction of distinct language features activate at each layer:

\begin{equation}
S_l = \frac{1}{d_{sae}} \left( \frac{1}{N} \sum_{i=1}^{N} ||\text{SAE}_l(\mathbf{V}_l^i)||_0 \right)
\end{equation}

where $||\cdot||_0$ counts the number of nonzero elements and $d_{sae}$ is the dictionary size of the SAE. If sparsity levels for visual representations are comparable to those of language representations, then this suggests that the visual representations have mapped to the language model features.

To validate that the features activated for the visual representations are encoding visual information, we measure the alignment of the feature descriptions to the visual input. To do this we apply the SAE to all visual token positions in each layer. To avoid uninterpretable high-frequency features, those activating across a significant number of inputs, we remove features that activate on more than 5\% of our image dataset, or more than 0.5\% of the original LLM dataset used to create the feature descriptions. We then obtain the top three strongest activating SAE features across all visual tokens, and prompt GPT-4o to assess whether any of the feature descriptions for the three selected features strongly matches a concept in the image.

\subsection{Experimental Setup}

We apply this methodology, using 5000 examples from the LLaVA-1.5 finetuning dataset to compute the reconstruction and sparsity errors. To measure the alignment of SAE feature descriptions and image content we use 1000 examples. For each example, consisting of an image, an instruction and an answer, we create a text-only example using "Consider the following information: " followed by the answer (containing the visual information that is asked for), followed by the question. We obtain activations for the text-only input by running them through Gemma-2-2b-it. We use these activations as a baseline for each SAE experiment.

We run the reconstruction and sparsity experiment across all layers and across the representations at the image token positions in the VLM, the text token positions in the VLM and all token positions for the text-only baseline examples (excluding the BOS token). For the description alignment experiment we use the same representations except those at the VLM text token positions, as they trivially contain visual features that stem from the question given in text format, and not from the image, and are therefore not interesting for this experiment.

\subsection{Results and Findings}

\begin{figure}[t]
    \centering
    \includegraphics[width=\linewidth]{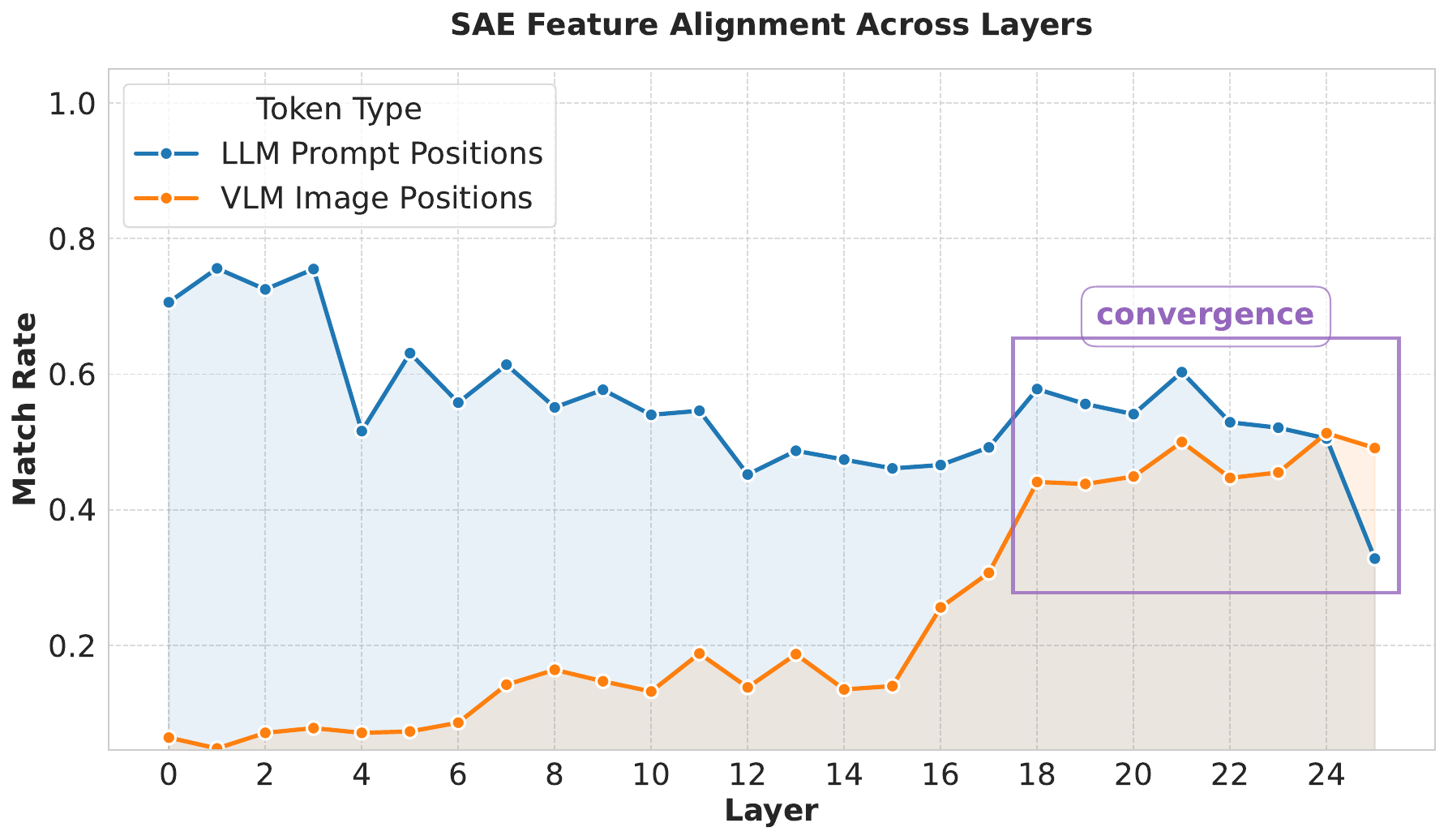}
    \caption{Semantic alignment of SAE feature descriptions across model layers. The y-axis shows the rate with which a visual SAE feature was found for each example. The x-axis shows the layer index.}
    \label{fig:sae_alignment}
\end{figure}

The results, depicted in figure \ref{fig:reconsparsity} and \ref{fig:sae_alignment}, show that the reconstruction error, sparsity, and semantic alignment all converge around layer 18 (out of 26) in the Gemma-2-2b-it model. This indicates that visual representations are fully mapped onto LLM features in these layers while remaining misaligned in earlier layers. These findings highlight a structured cross-modal integration process, where visual features transition into meaningful language model representations in specific mid-to-late layers. Future work should explore cross-modal alignment in end-to-end fine-tuned models and how to encourage earlier alignment, potentially improving learning efficiency.

\section{Conclusion}
This work presented a methodological framework for analyzing the mapping of visual features to language representations in multimodal LLMs. Following the LiMBeR framework \cite{merullo2023linearly}, we connect a frozen vision encoder to a frozen language model through a trainable projector, to force the adapter to learn to translate visual information into the language model's existing semantic space. 
Then we leverage pre-trained sparse autoencoders for the frozen LLM to track how visual information flows through the language model, identifying where vision tokens become aligned with language features. 
Our reconstruction error analysis and semantic feature investigation revealed how the connector bridges modalities, demonstrating when and how visual information is mapped to language semantics.

\newpage
{
    \small
    \bibliographystyle{ieeenat_fullname}
    \bibliography{main}
}


\end{document}